\documentclass[conference]{IEEEtran}
\IEEEoverridecommandlockouts
\usepackage{cite}
\usepackage{amssymb}
\usepackage{bbm}
\usepackage{amsmath,amsfonts}
\usepackage{algorithmic}
\usepackage{graphicx}
\usepackage{soul}
\usepackage{textcomp}
\usepackage{xcolor}
\usepackage{threeparttable}
\def\BibTeX{{\rm B\kern-.05em{\sc i\kern-.025em b}\kern-.08em
    T\kern-.1667em\lower.7ex\hbox{E}\kern-.125emX}}
\begin{document}

\title{Semantic Segmentation of Fruits on Multi-sensor Fused Data in Natural Orchards\\
\thanks{Identify applicable funding agency here. If none, delete this.}
}

\author{\IEEEauthorblockN{Hanwen Kang$^{*, \dagger}$}
\IEEEauthorblockA{\textit{College of Engineering} \\
\textit{South China Agricultural University}\\
Guangzhou, China\\}
\\
$^*$ Corresponding author \\
$^\dagger$ These authors contributed equally \\

\and
\IEEEauthorblockN{Xing Wang$^{a,b,\dagger}$}
\IEEEauthorblockA{\textit{$^a$Mechanical and Aerospace Engineering} \\
\textit{Monash University}\\
Melbourne, Australia \\}
\IEEEauthorblockA{\textit{$^b$Robotics and Autonomous Systems Group} \\
\textit{Data61, CSIRO}\\
Brisbane, Australia\\}
}

\maketitle

\begin{abstract}
Semantic segmentation is a fundamental task for agricultural robots to understand the surrounding environments in natural orchards. The recent development of the LiDAR techniques enables the robot to acquire accurate range measurements of the view in the unstructured orchards. Compared to RGB images, 3D point clouds have geometrical properties. By combining the LiDAR and camera, rich information on geometries and textures can be obtained. In this work, we propose a deep-learning-based segmentation method to perform accurate semantic segmentation on fused data from a LiDAR-Camera visual sensor. Two critical problems are explored and solved in this work. The first one is how to efficiently fused the texture and geometrical features from multi-sensor data. The second one is how to efficiently train the 3D segmentation network under severely imbalance class conditions. Moreover, an implementation of 3D segmentation in orchards including LiDAR-Camera data fusion, data collection and labelling, network training, and model inference is introduced in detail. In the experiment, we comprehensively analyze the network setup when dealing with highly unstructured and noisy point clouds acquired from an apple orchard. Overall, our proposed method achieves $86.2\%$ mIoU on the segmentation of fruits on the high-resolution point cloud (100k-200k points). The experiment results show that the proposed method can perform accurate segmentation in real orchard environments.
\end{abstract}


\section{Introduction}
With the recent development of the LiDAR techniques, data in 3D point cloud format has been gradually used in robotics tasks under agricultural scenarios because it is robust to the variances in lighting, shadows, and other factors. The 3D point cloud can provide accurate and robust range measurements of the real world. Additionally, RGB cameras can still provide dense texture information missing from LiDAR data, which can cooperate with Lidar to receive rich information \cite{wang2022geometry}. With proper extrinsic calibration of the cameras and LiDARs, the point cloud with texture information of the real world can be obtained. Semantic scene understanding is a fundamental task in many agricultural scenarios, such as growth monitoring and robotics perception tasks \cite{zhou2022intelligent,westling2021procedure,scalisi2021reliability}. However, semantic segmentation of 3D point clouds, particularly segmentation of large-scale point clouds, is still a challenging task as 3D point clouds are generally unstructured and unordered \cite{qi2017pointnet,zhuang2021perception}. How to improve the performance of semantic segmentation of 3D point cloud data in these agricultural scenes has attracted more and more attention.

Semantic segmentation plays a core role in computer vision. It groups image pixels together which belong to the same class, which is visually meaningful areas for analysis and understanding \cite{wang2018understanding}. Semantic segmentation has been widely used in robotic perception, medical image processing, autonomous driving, agricultural harvesting etc \cite{valada2017adapnet,asgari2021deep,feng2020deep,lin2021three }.
The recent development in Convolutional Neural Networks (CNN) provides remarkable performance improvement as compared with the traditional classifiers like random forest \cite{yamashita2018convolutional}. 
Jonathon et al. \cite{long2015fully} developed a fully convolutional network (FCN) for end-to-end image segmentation, which achieved a 20\% increase in the segmentation performance on Pascal VOC 2012 database. 
Olaf et al. \cite{ronneberger2015u} proposed a UNet network and training strategy that highly utilizes the data augmentation with available annotated samples for biomedical image segmentation. 
The authors further developed the Deeplab-v3 network and applied ResNet/MobileNet as the backbone \cite{chen2017rethinking}. Modules with dilate convolution in the cascade were designed to capture multi-scale context by adopting multiple dilate rates. Additionally, the augments on dilate Spatial Pyramid Pooling was implemented to probe convolutional features at multiple scales. 

The semantic segmentation method has also been widely applied to promote the advancement of the agricultural field. Zou et al. proposed a UNet-based algorithm to segment weeds from similar crops under a field environment \cite{zou2021modified}. The segmentation network was trained using a two-stage training method, including pre-processing and fine-tuning, which achieved an IOU of 92.92\%.
Wu et al. applied DeepLabV3 models with four backbones of ResNet-50, ResNet-101, Xception-65, and Xception-71 to segment abnormal leaves of hydroponic lettuce \cite{wu2021segmentation}. ResNet-101 had the best segmentation performance in the uniform weight assignation method with pixel accuracy of 99.2\%, and mIoU of 0.8326. Peng et al. applied FCN to segment the dense grapes with different varieties, and an IOU of 75.61\% was achieved on the RGB images. He also applied UNet and DeeplabV3 to compare the segmentation accuracy, which ended up with an accuracy of 77.53\% and 84.26\% for the IOU \cite{peng2021comparative}. 

Even though the 2D semantic segmentation achieves compatible segmentation accuracy and robustness, its performance experiences significant degeneration while dealing with complex natural orchards, especially for highly occluded large-scale scenes. Additionally, purely 2D images can not describe the 3D space well due to a lack of depth information. Recent research has focused on the semantic segmentation of 3D point clouds. 
Wei et al. proposed BushNet to achieve the semantic segmentation of 3D points in a large-scale agroforestry environment \cite{wei2022bushnet}. It included a minimum probability of random sampling module that can quickly and randomly sample a huge point and a multi-channel attention module to improve the attention distribution accuracy and training efficiency. The segmentation accuracy was improved by 12\% with the additional modules implemented.  
Chen et al. introduced RandLa-Net for large-scale unstructured agricultural scenes \cite{chen20213d}. A local feature aggregation module was integrated and improved to achieve the large-scale 3D point cloud segmentation. The experiment results suggested the best segmentation accuracy of 94\%, and the mIoU can reach 74\%. 
Yu et al. designed LFPNet that can directly consume fruit point clouds in real scenes to deal with classification error, incomplete segmentation, and low-efficiency  \cite{yu2022lfpnet}. The final results achieved an average segmentation accuracy and mIOU of 80.2\% and 76.4\%, respectively. However, the data were collected from the facility environment with a Kinect-v2 RGB-D camera, which will experience high-performance degeneration under natural lighting conditions. The network can not be deployed to mobile devices or be used for on-site segmentation tasks.

Despite the significant importance of semantic segmentation on 3D point clouds, it has been rarely studied for scene segmentation in unstructured scenes of real orchards. The point clouds of the surrounding environment and other factors in the real orchard are highly unstructured, uneven distributed, and noisy even compared to the point clouds acquired in city scenes. These pose huge challenges in performing semantic segmentation directly on 3D point clouds.

This study proposes a PointNet-based 3D segmentation method to perform segmentation in natural orchards. Two critical challenges are solved in this work. The first challenge is how to efficiently fuse features from images and point clouds to perform semantic segmentation. Since image data can provide enriched texture information and point cloud data can provide geometrical information about the world. The second challenge is how to train the point-based semantic segmentation under imbalance class conditions efficiently. A complete point cloud data of an orchard scene may contain more than 100k points, while only a few belong to the foreground objects. Such problematic class distribution could severely affect the accuracy of the trained model. Yu et al.'s work \cite{yu2022lfpnet} is the most similar work to our method. However, their work does not discuss the multi-sensor data fusion and training under imbalance class, which are two critical problems of 3D semantic segmentation orchard scenes. Meanwhile, their method requires a high-precision multi-viewpoint cloud that can only be obtained in ideal environments. While Our method only requires the fused RGB-point data that can be collected by LiDAR-Camera in the orchard environments.

In this paper, we propose a deep-learning-based 3D segmentation network to segment fruits in the high-resolution (100k - 200k points) colorized point clouds. We leverage the PointNet++ architecture \cite{qi2017pointnet++}, an existing SOTA deep-learning model in processing point clouds, and present an improved PointNet++ architecture that can fuse texture and geometrical information from a LiDAR-Camera visual sensor. We show that the accuracy of the PointNet-based model can be significantly improved by efficiently designing the fusion behaviour of the color and point clouds. To further promote the performance of the presented network, we utilize under-sampling and Weighted Cross-Entropy (WCE) to improve the network training efficacy. By distilling multiple network designs and training strategies, we obtain a comparatively accurate and robust 3D semantic segmentation method. Moreover, a complete implementation of the methods in terms of the LiDAR-Camera data fusion, data labelling and pre-processing, network training and inference is also introduced in detail. Our primary contributions to this paper are: 
\begin{itemize}
    \item A method that can perform semantic segmentation on high-resolution colorised point cloud data from a LiDAR-Camera fused sensor.
    \item A concise but efficient network architecture that can fuse features from image color and point cloud data.
    \item Demonstration of the proposed method from LiDAR-Camera fusion, and data labelling, to the network training and prediction, providing an end-to-end implementation for semantic segmentation in natural orchard environments. 
\end{itemize}

The rest of the paper is organized as follows. The methodologies of our approach are presented in Section \ref{section: method}. The experiment results and discussion are presented in Sections \ref{section: experiments} and \ref{section: discussion}, followed by the conclusion in Section \ref{section: conclusion}. 

\section{Materials and Methodology}\label{section: method}
\subsection{Method Overview}
\begin{figure*}[ht]
\centering
\includegraphics[width=0.99\textwidth]{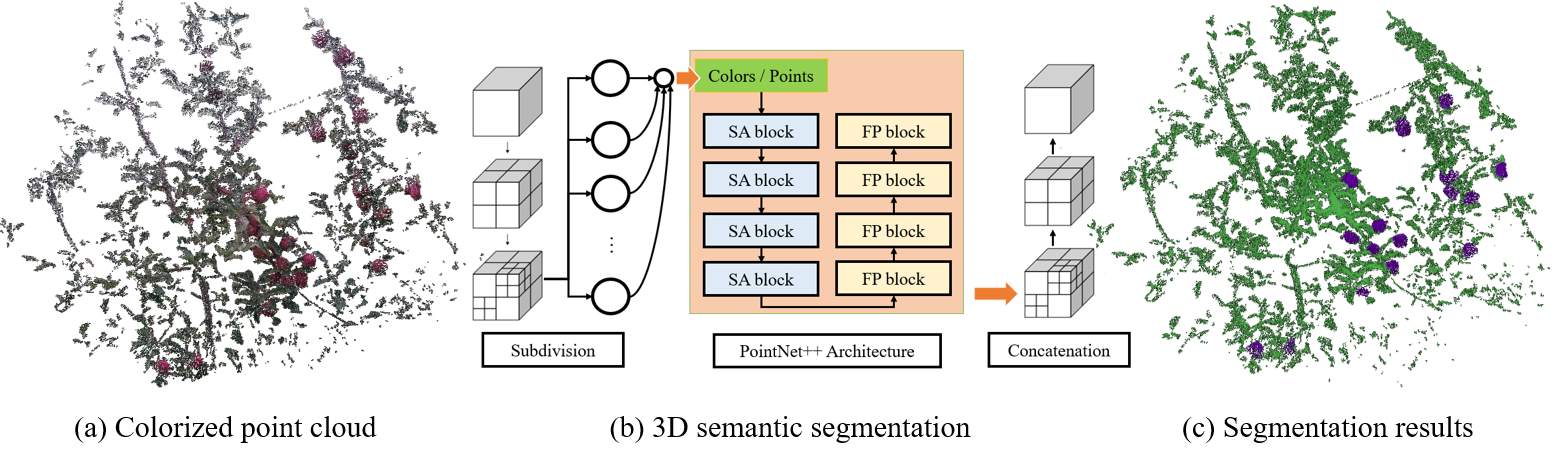}
\caption{Illustration of our proposed 3D semantic segmentation method. (a) the input data is colorized point clouds collected by using LiDAR-Camera, (b) the input point clouds are segmented by using the proposed segmentation module, which includes a spatial subdivision algorithm and a PointNet-based segmentation network, (c) the segmentation results.}
\label{fig: overview}
\end{figure*}
An overview of the proposed segmentation method is shown in Figure \ref{fig: overview}. The system captures sensor data from a Solid State LiDAR (SSL) Livox Mid-70 or Livox AVIA and an RGB camera. The intrinsic of the camera and extrinsic between LiDAR and camera are calibrated using the method from our previous work \cite{kang2022accurate}. We seek to estimate the semantic label of points that belong to the fruits using either point cloud or fused color information. This estimation problem can be formulated as a dense semantic segmentation problem. We leverage the PointNet++ architecture to perform semantic segmentation on input data since it shows SOTA performance in processing point cloud data in many core vision perception tasks. Our method can perform semantic segmentation on raw point cloud data and point cloud with color information. Note that without loss of generality, the proposed method can also incorporate measurements from other 3D range sensors, like spinning LiDARs. Mean Intersection over Union (mIoU) is used to evaluate the performance of the network model.

\subsubsection{Input Data}
\begin{figure}[ht]
\centering
\includegraphics[width=0.47\textwidth]{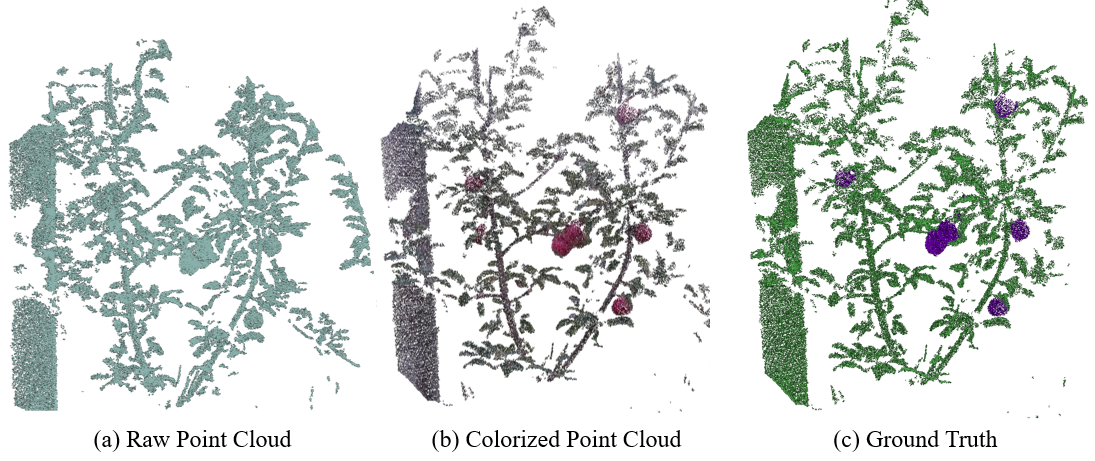}
\caption{The proposed segmentation method can use both (a) raw point cloud, (b) colorized point cloud. (c) is the an example of segmentation result.}
\label{fig: input data}
\end{figure}

Our method uses a set of sensors to collect data from the orchard environment. The inputs of the proposed method include point clouds and 2D color images. The Livox Mid-70 or Livox AVIA measures the surrounding geometry by sweeping over the scene with a set of lasers (one line for Livox Mid-70 and six lines for Livox AVIA). The data collection frequency of the Livox Lidar is 10 HZ. For Livox Mid-70, the LiDAR-camera sensor typically requires 10 seconds to obtain sufficient points to cover one single scene. In comparison, Livox AVIA only needs 3 seconds since it has more laser channels than the former. The input points are firstly fused with color information and then added to construct a complete point cloud for the current scene. A typical complete point cloud for a working scene (the width and height of the scene are 2.5 to 3 meters, the depth of the scene is between 1.2 to 2.5 meters) normally includes 100k - 200k points after post-processing. The post-processing steps comprise outlier rejection and voxelization. The size of the voxel grid is set as $1cm \times 1cm \times 1cm$ to preserve the details of the local-scale geometries. An example of the raw and colorized point cloud and semantic segmentation result of the scene is shown in Figure \ref{fig: input data}.

\subsubsection{Partitioned Inference on Point Cloud}
\begin{figure}[ht]
\centering
\includegraphics[width=0.48\textwidth]{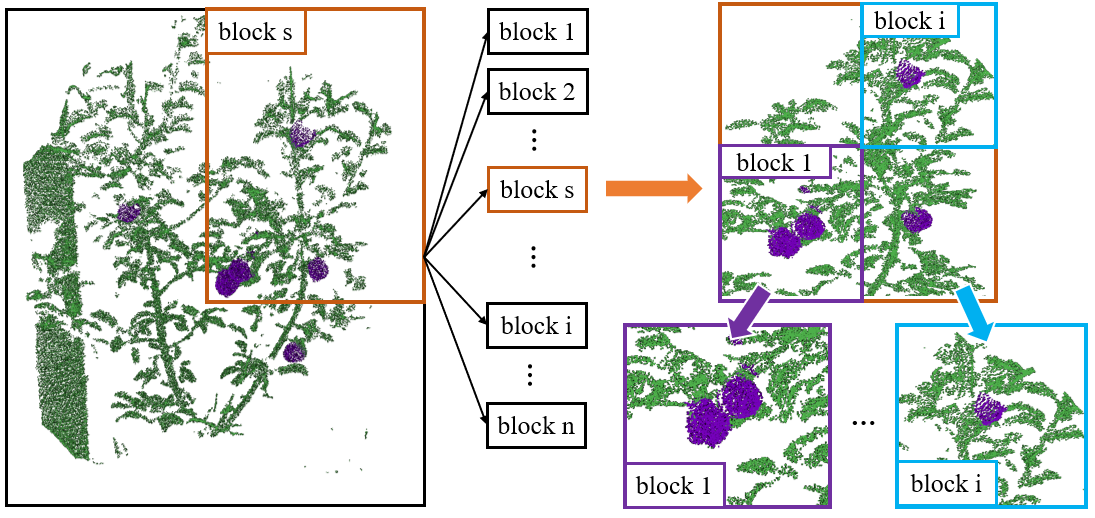}
\caption{Illustration of spatial subdivision of proposed method in segmentation inference. The complete scene is subdivided by an Octree until the point number within a leaf node is less than the given number, which is set as 4096 or 8192.}
\label{fig: subdivision}
\end{figure}

\begin{figure*}[ht]
\centering 
\includegraphics[width=0.92\textwidth]{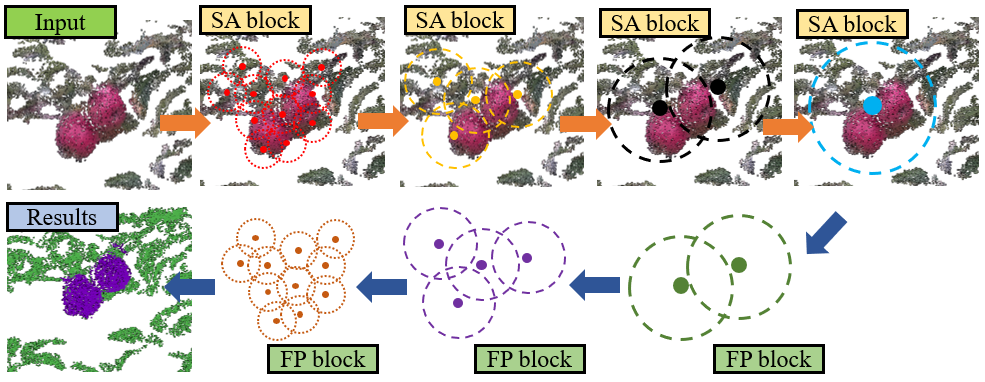}
\caption{PointNet++ uses a set of SA blocks and FP blocks to extract and process features from the point cloud. SA blocks under-sampling the points and extracts features from the point set by a local-to-global behaviour by increasing the grouping region of each chosen centroid. FP blocks up-sampling the points and propagates the processed features to each point within the set.}
\label{fig: pointnet}
\end{figure*}
Since the point cloud of a complete scene contains more than 100k points, which cannot be processed at once. Therefore, we partition the whole point cloud of a scene into multiple blocks through the Octree. Each node of the Octree should contain the points that are not beyond the given number. The typical value of the given number is 4096 and 8192, based on different situations. Based on the value of the given number of points, we divide the segmentation into two cases: small-scale segmentation and large-scale segmentation. Small-scale segmentation stands for performing segmentation on a node that contains a number of points less than 4096. In contrast, large-scale segmentation stands for performing segmentation on the node containing either 4096 or 8192 points. Additionally, fruits always count only a small part of points in the whole point cloud. Hence performing segmentation on the large-scale point cloud is more challenging due to the presence of an imbalanced distribution of classes. Therefore, finding a suitable point number value in each node is also a critical task to secure segmentation accuracy. During the forward inference, the complete point cloud is partitioned into small blocks and fed into the network one by one until all blocks are processed. Then, the predictions of these blocks are assembled to obtain the final results. The overall procedures of the partitioned inference as shown in Figure \ref{fig: subdivision}. 

\subsection{Network Architecture}\label{subsection: method-pts}
\subsubsection{PointNet++ Architecture}
PointNet++ network uses a hierarchical feature learning strategy to extract and learn features from the point cloud, as shown in Figure \ref{fig: pointnet}. It builds a hierarchical grouping of points, which can progressively learn features from larger and larger local regions along the hierarchy and aggregate local and global information. However, since the point cloud is unordered, highly non-uniform, and does not have a structured neighbourhood region similar to the image. Hence the Set Abstraction (SA) layer and Feature Propagation (FP) layer are used to process the point cloud in each hierarchical level and perform dense prediction by propagating point features along the neighbour region, respectively. The base PointNet++ network has four SA layers and four FP layers to process and predict the label of each point.

The SA layer comprises three layers: the sampling layer, grouping layer, and PointNet layer. Sampling layer sample a given number of centroids from the point cloud using the iterative farthest point sampling. Iterative Farthest Point Sampling (FPS) can ensure an even distribution of centroids within the point cloud. Then, the grouping layer is used to find the neighbour points within a local region of each centroid. Different grouping methods can be used, including K Nearest Neighbor (KNN) points or points in a certain radius. After that, the PointNet layer is used to extract features from each neighbour region of the centroids. With the increase of the hierarchical level, the SA layer gradually extracts the features from a small local area to a larger global region. In this manner, PointNet++ builds a hierarchical structure to learn multi-scale features from the point set.     

SA layer down-samples the points cloud by progressively increasing the local region size and reducing the number of centroids. Semantic segmentation is a dense classification task that requires labels on each point. Instead of using transpose convolution in a 2D segmentation network, PointNet++ uses the FP layer to propagate features from the centroids to the points of the local region. That is, the features of a point are from the interpolation of nearby centroids weighted based on the distance from the points to each centroid. The interpolated features of points are also concatenated with the skip-linked point features from the SA layer. Then the concatenated features are passed through a one-by-one convolution in CNNs and ReLu layers.

\begin{figure}[ht]
\centering
\includegraphics[width=0.45 \textwidth]{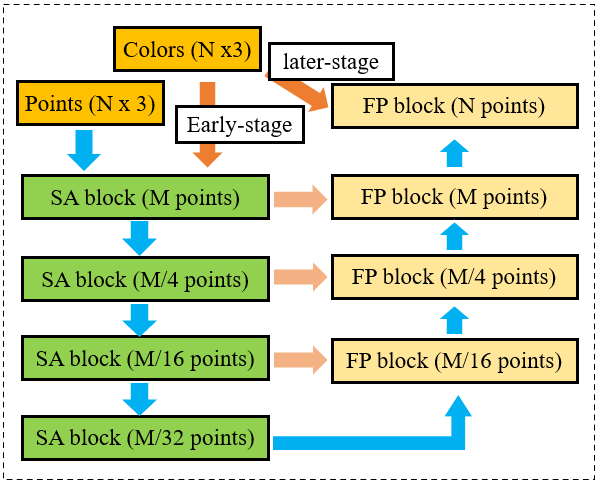}
\caption{The network architecture of the PointNet++ model, which includes four SA blocks and four FP blocks. color information of point cloud can be fused by either early-stage fusion or late-stage fusion ways.}
\label{fig: fusion model}
\end{figure}
\subsubsection{Fusion with color Feature}
The PointNet++ architecture can fuse points and other features by increasing the number of input data channels. For example, PointNet++ use the normal vector of each point within the point set in the classification of ModelNet40 dataset, and a $1.2\%$ improvement of accuracy is achieved. PointNet++ has fused the features at the beginning and the end of the network, named early-stage fusion and later-stage fusion, respectively, as shown in Figure \ref{fig: fusion model}. Similarly, we input XYZ and RGB values into the network together to test their effect on the segmentation performance. Unexpectedly, the segmentation accuracy of using the fusion data drops to 0.72, while the point-only network model can achieve 0.84, as listed in Table \ref{table: method comparison}. To investigate the reason why fusion data affect the segmentation accuracy, we respectively input color features in only early-stage fusion or later-stage fusion. Results show that later-stage fusion can improve the segmentation accuracy while the accuracy of early-stage fusion is reduced. 

\begin{figure}[ht]
\centering
\includegraphics[width=0.45 \textwidth]{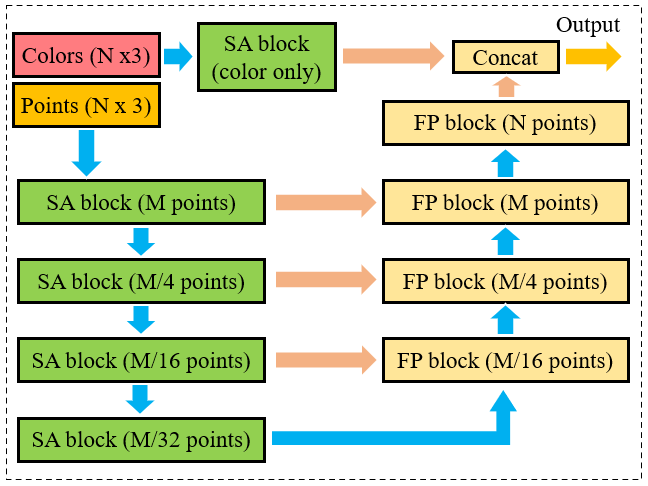}
\caption{The network architecture of the proposed Fuse-PointNet++ model. A SA block is used to extract color features of each point, and the processed feature is fused through the later-stage fusion.}
\label{fig: fuse pointnet}
\end{figure}
A possible reason leading to the accuracy degeneration introduced by early-stage feature fusion is the large difference between feature distributions. Since PointNet++ fuse the multi-sensor features by simply concatenating with each other and processing them with convolution operations. When those features have large differences in the distribution in feature space, this operation will affect the learning performance of the network on multi-sensor features. Therefore, we only introduce color features in later-stage fusion. Rather than directly using the RGB values on each point, we use a color-only SA block to gather color features from the neighbourhood of each point. The introduced later-stage fusion PointNet++ architecture is named as Fuse-PointNet++ model, and it achieves 0.88 on segmentation accuracy, as shown in Table \ref{table: method comparison}. The network architecture of the Fuse-PointNet++ model is shown in Figure \ref{fig: fuse pointnet}.

\subsubsection{Sampling and Grouping Strategy}
PointNet++ can apply FPS or Random Sampling (RS) strategy to sample $N$ centroids from a point cloud with the number of $M$ points and then uses KNN or distance to gather $K$ points within the neighbourhood of each centroid. We use the FPS method as it can sample evenly distributed centroids from the point cloud. Then, a given number of points within a certain radius is gathered to extract the local features of each centroid. 

\subsection{Class Imbalance Training}
\subsubsection{Training Method}
Class imbalance problem occurs when one class has much more samples than other classes, which is one of the most significant issues in network training \cite{yuan2018regularized}. In our case, a point cloud can have up to 16k points, while only a small subset of points (less than 1k) belongs to positive samples. The imbalance class can cause training degeneration since most samples are easy negatives that contribute no useful backward gradient. Introducing a weighting factor on training loss or Under-Sampling (US) are the two common methods to deal with the class imbalance problem. Therefore, we introduce the under-sampling strategy for training data pre-processing and introduce a weighting factor on loss function to alleviate the training degeneration caused by the imbalance class problem. 
\begin{figure}[ht]
\centering
\includegraphics[width=0.43 \textwidth]{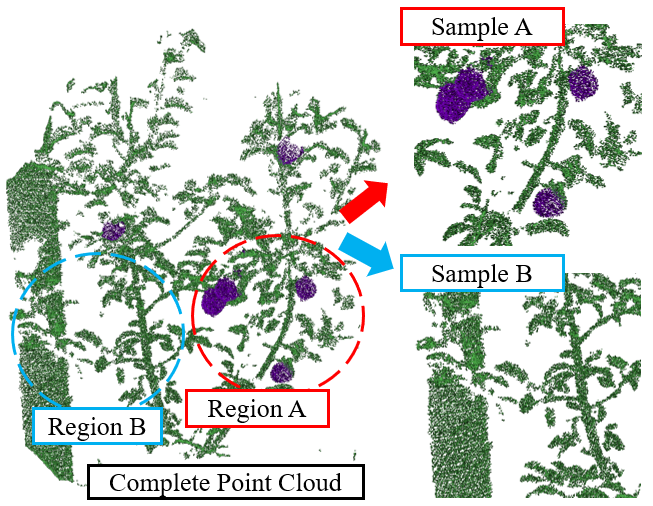}
\caption{(a) Region A has sufficient fruit points while (b) region B has only background points. Region A is used in network training while region B is removed.}
\label{fig: under sampling}
\end{figure}

\begin{figure*}[ht]
\centering
\includegraphics[width=0.97 \textwidth]{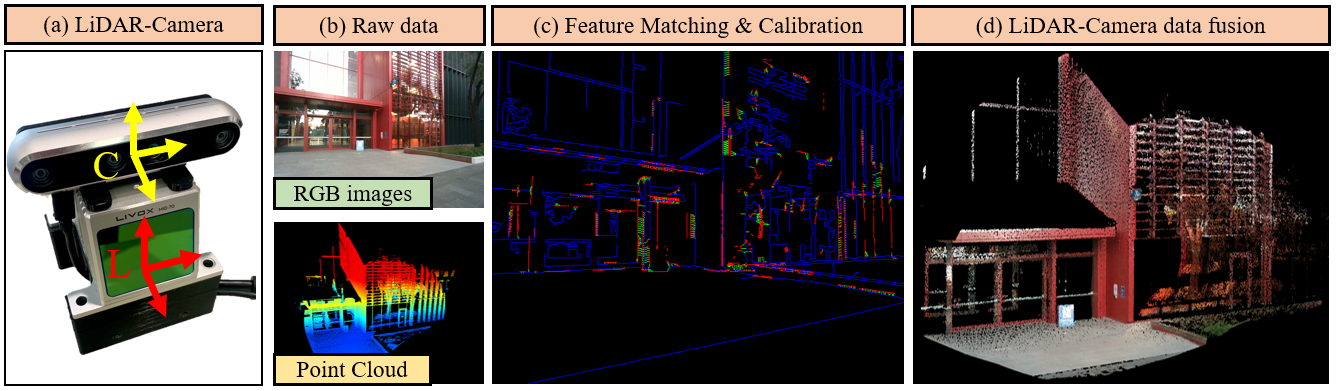}
\caption{A illustration of the used LiDAR-Camera visual sensor, (a) it includes a RealSense-D455 and a Livox Mid-70;  (b) the raw input of the visual sensor are RGB images from the camera and the point cloud from the LiDAR; (c) LiDAR and camera can be automatically calibrated by performing feature extraction and matching \cite{yuan2021pixel}; (d) a demonstration of fused colorized point cloud.}
\label{fig: lidar system}
\end{figure*}
Firstly, we set the minimum number of fruits point within each training sample. That is, the training samples in the small-scale dataset (point number is 4096) should keep the points belonging to fruit larger than 1k (as shown in Figure \ref{fig: under sampling}), while the training samples in the large-scale dataset (point number is 8192) should have fruit points larger than 1.5k, respectively. 

Secondly, we introduce the weighting factor $\alpha \in [0, 1]$ to balance the loss value of the two classes. The loss function $\mathcal{L}(p)$ is formulated as follows:

\begin{equation}
    \mathcal{L}(p) = -\frac{1}{N}\sum_{c=0}^{M} \alpha_c \sum_{i}^{N} \mathbbm{1}_{c} y_i \mathbf{log}(p_i)
\end{equation}
where $y_i$ and $p_i$ are the label and prediction of the points, respectively. $\alpha_c$ is the weight factor of the points in class $c$. $\mathbbm{1}_{c}$ denotes the sign function whether points in class $c$ ($\mathbbm{1}_{c}$ = 1) or not ($\mathbbm{1}_{c}$ = 0). We denote the weighted cross entropy loss function as WCE in the following experiments. The preliminary results as shown in Table \ref{table: method comparison} show that US and WCE can improve the network training performance. A more detail experiments of parameter setup of US and WCE are introduced in the experiment.
\begin{table}[ht]
    \centering
    \caption{Comparison of network performance under different fusion and training strategies.}
    \begin{tabular}{c c c c}
    \hline
    Input & Fusion Method & Training Method & mIoU \\
    \hline
    pc & - & - & 0.73 \\
    pc & - & US & 0.79 \\
    pc & - & US + WCE & 0.81 \\
    pc + rgb & early+later fusion & US + WCE & 0.75 \\
    pc + rgb & early fusion & US + WCE & 0.72 \\
    pc + rgb & later fusion & US + WCE & 0.84 \\
    pc + rgb (SA) & later fusion & US + WCE & \textbf{0.88} \\
    \hline
    \end{tabular}
    \label{table: method comparison}
\end{table}

\subsubsection{Training Details}
During the training process, the learning rate and decay rate are respectively set as 0.001 and 0.95 per epoch, and the minimum learning rate is specified as 0.0001. The Adam optimizer is used to train the network, and the batch size is set from 4 to 16 based on the different number of input points (batch sizes are 4 for 8192, and 16 for 4096, respectively) due to limited memory consumption on GPU. Data augmentation is conducted on both points and color images. For point set data, augmentations such as rotation and adding white noise are introduced, while adjustments of saturation and brightness are introduced on color information. We train the network model for 100 epochs and use the checkpoint with the best validation accuracy in the last epochs as the final network weights.

\subsection{System Implementations}
\subsubsection{LiDAR-camera Fusion}
To build a customized and large 3D dataset, we use the LiDAR-Camera sensors that include a pin-hole camera model to acquire color images and a LiDAR for 3D points acquisition, as shown in Figure \ref{fig: lidar system}. The pin-hole camera model project a point $^{C}p_{i} \in R^{3}$ from the LiDAR coordinate to a pixel $^{C}\hat{p}_{uv} \in R^{2}$ on color image plane by using the equation:
\begin{equation}
    ^{C}\hat{p}_{uv} = \pi(K\hat{T}^{C}_{L}\ ^{L}p_{i})
\end{equation}
where $\pi( \cdot )$ is the camera distortions model, $K$ is the intrinsic matrix of camera, $\hat{T}^{C}_{L} = (\hat{R}^{C}_{L},\hat{t}^{C}_{C,L}) \in SE(3)$ (where $\hat{R}^{C}_{L} \in SO(3)$ and $\hat{t}^{C}_{C,L} \in R^{3}$) is the extrinsic between the LiDAR and camera. The intrinsic of the  camera is evaluated firstly, and then the calibration between LiDAR and camera can be conducted in either manual or automatic way.

\subsubsection{Data Collection and Labelling}
The data were collected in the Fankhauser apple orchard located in Melbourne, Australia, using the proposed Lidar-camera sensor. The collection time varied from 10:00 am to 4:00 pm. The distance between the vision sensor to the tree trunk ranged from 1.2 to 2.5 meters. The scanning height was adjusted to guarantee the apple trees are within the FoV of both sensors. In total, there were 152 point cloud data pairs, including ROSbag or PCD format and 152 corresponding color images. The point clouds were recorded by setting the scanning time to 10 seconds to accumulate sufficient points of the scene. Each point cloud contains approximately 100k to 200k points. The ground truth of the semantic segmentation was labelled using the open-source labelling tool on Github $^{1}$. 
\footnotetext[1]{https://github.com/Hitachi-Automotive-And-Industry-Lab/semantic-segmentation-editor}

\subsubsection{Hardware and Software}
Both the LiDARs and RGB camera (Intel Realsense D-435 or D-455) are connected with an Nvidia Xavier using the Robotic Operation System (ROS) in the Melodic version of Ubuntu 18.04. The data transmission between the Livox LiDAR and camera and Nvidia Xavier is through the Ethernet and USB port. The communication between Livox LiDARs and Intel RealSense cameras is achieved by using the Livox-ROS-Driver and Realsense-ROS-SDK, respectively. Other required packages such as LiDAR-RGB calibration and fusion nodes are also programmed and running on ROS. The network models are developed based on Tensorflow-1.15. The network training is performed using an NVIDIA RTX-3060 (6GB), and the forward inference is tested on both NVIDIA RTX-3060 and NVIDIA Xavier.

\section{Experiments}\label{section: experiments}
\subsection{Experiment Method}
In this section, we evaluate the proposed methods in four experiments. Firstly, an ablation study of the network model on small-scale and large-scale scenes is presented. Then, the experiment of semantic segmentation on a complete point cloud is conducted. After that, the comparison studies of 3D and 2D segmentation are presented, and segmentation on data from RGB-D cameras is presented. In the experiment, each network is trained three times, and the network weights of the best validation accuracy are saved for performance evaluation. In the experiments, mIoU is used to evaluate the segmentation performance of the network.

We use 76 of 152 scenes to create the dataset for network training, and the rest of the scenes are used to perform training validation and evaluation. We randomly choose 100-200 seeds from each point cloud and create the training dataset by crop the neighbour points of each seed. The training dataset is classified into small-scale ($M=4096$) and large-scale ($M=8192$) datasets. Each dataset includes from 10000-15000 training samples by using the proposed pre-processing methods.

\subsection{Ablation Study of Model}
\begin{table}[ht]
    \centering
    \caption{Comparison of sampling and grouping strategies on network performance}
    \begin{tabular}{c c c c c}
    \hline
    index & Method & $N$ & $K$ & mIoU \\
    \hline
    1 & KNN  & 1024 & 24 & 0.868 \\
    2  & \textbf{FPS}  & \textbf{1024} & \textbf{24} & \textbf{0.872} \\
    3  & FPS  & 1024 & 12 & 0.842 \\
    4  & FPS  & 1024 & 48 & 0.857 \\
    5  & FPS  & 512 & 24 & 0.833 \\
    6  & FPS  & 2048 & 24 & 0.867 \\
    \hline
    \end{tabular}
    \label{table: sampling and grouping}
\end{table}
We firstly analyse the effect of network parameters on segmentation performance. Table \ref{table: sampling and grouping} shows the evaluation of segmentation accuracy on centroids sampling and neighbour points grouping strategy by using the small-scale dataset ($M$=4096) on the Fuse-PointNet++ model. Tests 1 and 2 compare the model that uses KNN and FPS sampling strategies, respectively. Experimental results show that FPS and KNN do not affect the accuracy significantly, while the network with FPS achieves a slightly higher score compared to the method of using KNN. Tests 2, 3, and 4 compare the network performance of using a different number of grouping points. Results show that the network achieves the best performance when $K=24$. Tests 2, 5, and 6 compare the network performance o using a different sampling centroid. Results show that under-numbered sampling centroids would reduce the accuracy of the segmentation, while over-number of sampling centroids would not improve the segmentation accuracy but the computational consumption is significantly increased.

\begin{table}[ht]
    \centering
    \caption{Comparison of grouping Scale on network performance}
    \begin{tabular}{c c c c}
    \hline
    index & Scale & Nor (Scale) & mIoU \\
    \hline
    1 & 0.1 $\hookrightarrow 0.2 \hookrightarrow 0.4 \hookrightarrow 0.8$ & False & 0.813 \\
    2 & 0.05 $\hookrightarrow 0.1 \hookrightarrow 0.2 \hookrightarrow 0.4$ & False & 0.836 \\
    3 & 0.01 $\hookrightarrow 0.02 \hookrightarrow 0.04 \hookrightarrow 0.08$ & \textbf{False} & \textbf{0.872} \\
    4 & 0.1 $\hookrightarrow 0.2 \hookrightarrow 0.4 \hookrightarrow 0.8$ & True & 0.742 \\
    5 & 0.05 $\hookrightarrow 0.1 \hookrightarrow 0.2 \hookrightarrow 0.4$ & True & 0.746 \\
    6 & 0.01 $\hookrightarrow 0.025 \hookrightarrow 0.05 \hookrightarrow 0.1$ & True & 0.752 \\
    \hline
    \end{tabular}
    \label{table: grouping Scale}
\end{table}

Table \ref{table: grouping Scale} shows the ablation experiment using a different scale of neighbour points grouping by using the Fuse-PointNet++ model on the small-scale dataset. The results of tests 1, 2, and 3 show different grouping scales can lead to different segmentation accuracy. A small grouping scale achieves a higher accuracy on segmentation results. Tests 4, 5, and 6 show the network performance by normalizing the scale of the input point cloud. Results show that the segmentation accuracy of the network with scale normalizing drops compared to the un-normalized case. Since fruits always have a physical scale in the real world, a proper grouping scale can help the network to learn local features efficiently. Meanwhile, since input clouds always have different spatial sizes and scale normalization changes the scale of the data, hence the network performance on segmentation is reduced.

\begin{table}[ht]
    \centering
    \caption{Comparison of color fusion on network performance}
    \begin{tabular}{c c c c}
    \hline
    Index & Method & No. channels & mIoU \\
    \hline
    1 & pc & - & 0.812 \\
    2 & raw RGB & 3 & 0.841 \\
    3 & W/o grouping & 32 & 0.847 \\
    4 & grouping & 32 & 0.872 \\
    5 & grouping & 16 & 0.867 \\
    6 & \textbf{grouping} & \textbf{64} & \textbf{0.874} \\
    \hline
    \end{tabular}
    \label{table: color fusion}
\end{table}
Table \ref{table: color fusion} shows the experiment result of using the different setups on color fusion by using the Fuse-PointNet++ model on the small-scale dataset. Tests 1 to 4 evaluate the segmentation accuracy of the network by using the different fusion methods. Method 1 only uses the point cloud, method 2 concatenates raw RGB values to the feature vectors, method 3 uses a one-by-one convolution operation to process color information, and method 4 applies a SA block to gather the neighbour of each point and then process the color information within this neighbour region. Experimental results show that method 3 achieves the best accuracy on network performance. These results show that grouping can help the network to gather the spatial distribution of the color features, which is similar to the convolution operations in 2D image processing.  Tests 4, 5, and 6 further investigate the proper number of color channels for feature fusion. Experimental results show that although increasing feature channels can slightly improve the segmentation accuracy, it also brings higher computational requirements. Therefore, we choose the feature channel number 32 in this work. 

\begin{figure*}[ht]
\centering
\includegraphics[width=0.98 \textwidth]{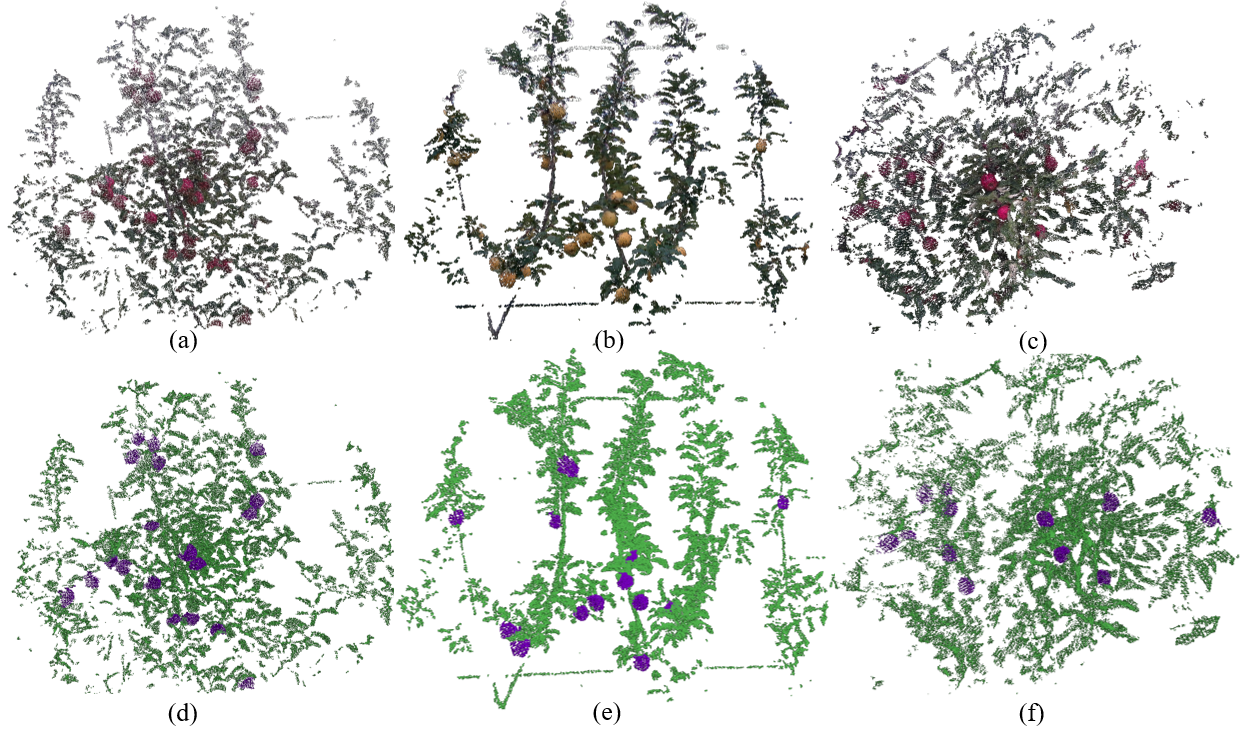}
\caption{The semantic segmentation of complete scene using the proposed method. (a), (b), and (c) are the input colorized point cloud, (d), (e), and (f) are the segmentation results of (a), (b), and (c), respectively.}
\label{fig: segmentation result}
\end{figure*}
From the results in this section, we set grouping points number $K$=24, grouping scale as $0.01 \hookrightarrow 0.02 \hookrightarrow 0.04 \hookrightarrow 0.08$, and the fusion color feature channels as 32. 

\subsection{Experiment on Training Strategy}
This section uses different training strategies in network training and evaluates the segmentation accuracy of the obtained model. We first evaluate the training method to find the optimal parameters for WCE on the small-scale dataset. In the small-scale training dataset, the average points number of fruits in a point cloud is 1200 (minimum fruits point in training samples is 1000) and the total point number of the point cloud is 4096. We use the optimal network setup from the last section and evaluate the effect of US and WCE strategies on trained network performance, the experimental results are shown in Table \ref{table: small-scale}.
\begin{table}[ht]
    \centering
    \caption{Comparison of training strategy on network performance in small-scale dataset}
    \begin{threeparttable}
    \begin{tabular}{c c c c c c}
    \hline
    Index & Method & No. pts & $\alpha_{nobj}$ & $\alpha_{obj}$ & mIoU \\
    \hline
    1 & CE & 250 & 1.0 & 1.0 & 0.757 \\
    2 & CE & 500 & 1.0 & 1.0 & 0.803 \\
    3 & WCE & 500 & 0.75 & 1.25 & 0.822 \\
    4 & CE & 1000 & 1.0 & 1.0 & 0.872 \\
    5 & WCE & 1000 & 0.85 & 1.15 & 0.877 \\
    6 & WCE & 1000 & \textbf{0.75} & \textbf{1.25} & \textbf{0.881} \\
    7 & WCE & 1000 & 0.65 & 1.35 & 0.865 \\
    \hline
    \end{tabular}
    \begin{tablenotes}
    \footnotesize
    \item[1] \textbf{CE} stands for Cross Entropy.
    \item[2] \textbf{No. pts} is the minimum fruit points in training samples.
    \end{tablenotes}
    \end{threeparttable}
    \label{table: small-scale}
\end{table}

The results of tests 1, 2, and 4 show that the imbalance class problem would severely affect the performance of the model, while under-sampling in the data pre-processing step could largely alleviate this training degeneration. After the US on training data (test-4), the segmentation accuracy of the model is significantly improved compared to the training result without US (test-1). Tests 2 and 3, 4 and 6 compare the network performance by respectively using cross-entropy loss and WCE. Results show that WCE can improve approximately $1\%$ to $2\%$ compared to the network trained by the cross-entropy. Tests 5 to 7 compare the network performance by using different weight factors in WCE. Results show that different values in WCE can also influence network performance. For the small-scale dataset, we set $\alpha_{nobj}$ and $\alpha_{obj}$ respectively as 0.75 and 1.25 since the network achieves the best accuracy when using these values.

We further evaluate the training strategies on a large ($M$=8192) dataset. The network setup follows the same setup as the small-scale dataset. The number of centroids points in the sampling step is set as 2048 and the training batch size is 6. In the middle-scale training dataset, the average points number of fruits in a point cloud is 1800 (minimum fruits point in training samples is 1500) and the total point number of the point cloud is 8192. The experiments are shown in Table \ref{table: large-scale}.
\begin{table}[ht]
    \centering
    \caption{Comparison of training strategy on network performance in large-scale dataset}
    \begin{threeparttable}
    \begin{tabular}{c c c c c c c}
    \hline
    Index & $N^{1}$ & Method & No. pts$^{2}$ & $\alpha_{nobj}$ & $\alpha_{obj}$ & mIoU \\
    \hline
    1 & 2048 & $CE^{3}$ & 500 & 1.0 & 1.0 & 0.667 \\
    2 & 2048 & CE & 1000 & 1.0 & 1.0 & 0.712 \\
    3 & 2048 & CE & 1500 & 1.0 & 1.0 & 0.767 \\
    4 & 1024 & CE & 1500 & 1.0 & 1.0 & 0.722 \\
    $5^{4}$ & 3072 & CE & 1500 & 1.0 & 1.0 & 0.784 \\
    6 & 2048 & WCE & 1500 & 0.75 & 1.25 & 0.786 \\
    7 & 2048 & WCE & 1500 & 0.60 & 1.40 & 0.797 \\
    8 & 2048 & \textbf{WCE} & \textbf{1500} & \textbf{0.50} & \textbf{1.50} & \textbf{0.808} \\
    9 & 2048 & WCE & 1500 & 0.40 & 1.60 & 0.793 \\
    \hline
    \end{tabular}
    \begin{tablenotes}
    \footnotesize
    \item[1] \textbf{N} is the centroid number in sampling step.
    \item[2] \textbf{No. pts} is the minimum fruit points in training samples.
    \item[3] \textbf{CE} stands for Cross Entropy.
    \item[4] Training batch size set as 3.
    \end{tablenotes}
    \end{threeparttable}
    \label{table: large-scale}
\end{table}

Tests 1, 2, and 3 evaluate the network performance by setting a different minimum number of fruit points in data pre-processing. Results show that imbalance class problems could lead to severe performance degeneration in large-scale dataset training. Increasing the minimum number of fruits point in training samples could improve the network performance to a certain degree. Tests 3, 4, and 5 evaluate the network performance in the large-scale dataset by using different centroids numbers in the sampling step. Results show that an increase of centroids can improve the final segmentation accuracy but the computational consumption also grows. Therefore, we choose the centroids number $N=2048$ as it achieves the balance between performance and computational efficiency. Tests 6 to 9 evaluate the network performance by using the WCE during the training process. Results show that WCE can improve the training results by re-weighting the loss. The proposed Network achieves the best accuracy (mIoU = 0.808) when $\alpha_{nobj}$ and $\alpha_{obj}$ are set as 0.5 and 1.5, respectively. 

\subsection{Segmentation in Orchards}
\begin{figure}[ht]
\centering
\includegraphics[width=0.47 \textwidth]{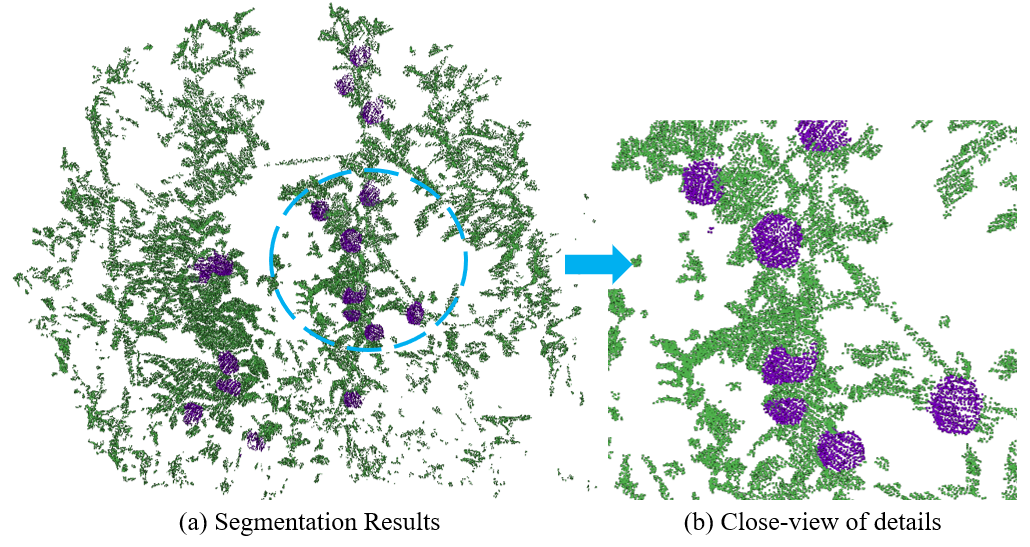}
\caption{Details of the semantic segmentation on point cloud using the proposed method.}
\label{fig: segmentation details}
\end{figure}
After evaluation of the key network parameters and training strategies respectively in the small-scale and large-scale datasets, this section evaluates the network performance on semantic segmentation of the complete point cloud. The input point cloud is firstly subdivided into small blocks by the Octree algorithm, each node should contain a point that less than the given threshold. We respectively use the network trained by using the small-scale dataset (network-S) and large-scale dataset (network-L). A complete point cloud of a scene after octree subdivision always contains 15 to 30 local blocks for network-L to process, and a double number of blocks for network-S to process. The accuracy of the network-S and network-L on complete scene segmentation is shown in Table \ref{table: complete scene}. 
\begin{table}[ht]
    \centering
    \caption{Accuracy evaluation of network on complete scene segmentation}
    \begin{threeparttable}
    \begin{tabular}{c c c c}
    \hline
    Index & Method & No. blocks$^{1}$ & mIoU \\
    \hline
    1 & Network-S & 28.8 & 0.862 \\
    2 & Network-L & 17.4 & 0.794 \\
    \hline
    \end{tabular}
    \begin{tablenotes}
    \footnotesize
     \item[1] No. blocks is the mean number of subdivided blocks of a scene.
    \end{tablenotes}
    \end{threeparttable}
    \label{table: complete scene}
\end{table}

The result shows that both networks trained by small-scale and large-scale datasets have a slight accuracy reduction on complete scene segmentation compared to their performance on the evaluation dataset. The possible reason is due to the different distribution of classes between training data and the original point cloud. As the utilized data pre-processing strategy under-sampling the non-objects classes to improve the training efficacy. Overall, the complete scene segmentation by using Network-S achieved $86.2\%$ on mIoU. The visualization results of the complete scene segmentation are shown in Figure \ref{fig: segmentation result}. Figure \ref{fig: segmentation details} shows the details of the segmentation result.

\subsection{Segmentation using RGB-D Cameras}
In our previous study, a LiDAR-Camera sensor and a multi-sensor data fusion method are developed to obtain high accuracy colorized point cloud from the orchard environments. In this work, we demonstrate a pipeline and a deep-learning-based network to perform semantic segmentation on the colorized point cloud. For comparison, we evaluate the performance of the proposed network with or without RGB information on data acquired by using the Intel RealSense-D455, a stereo-depth camera that has been widely applied in many agricultural types of research. We collect the RGB-D data of a tree from different distances (0.8m and 2.0m) in orchard environments, the resolution of the color and depth images are 640 $\times$ 480. The accuracy of semantic segmentation by using Network-S on RGB-D data from Intel RealSense-D455 is shown in Table \ref{table: rgbd}.
\begin{table}[ht]
    \centering
    \caption{Accuracy of network using the RGB-D camera}
    \begin{threeparttable}
    \begin{tabular}{c c c c}
    \hline
    Index & Distance & mIoU$_{pc}$ & mIoU$_{fused}$ \\
    \hline
    1 & 0.8 & 0.763 & 0.797 \\
    2 & 1.2 & 0.712 & 0.752 \\
    3 & 1.6 & 0.654 & 0.704 \\
    4 & 2.0 & 0.524 & 0.627 \\
    \hline
    \end{tabular}
    \end{threeparttable}
    \label{table: rgbd}
\end{table}

Results show that the accuracy of the segmentation drops dramatically with the increase in the distance between the sensor and the tree. Also, the network using fused data shows better accuracy compared to the network that only uses the point cloud, which shows that color information can help improve the accuracy of the segmentation when the point cloud has limited quality. These results also indicate that network using fused data has better robustness to deal with data quality degeneration in real implementations.

\subsection{Comparison with 2D segmentation method}
\begin{figure}[ht]
\centering
\includegraphics[width=0.49 \textwidth]{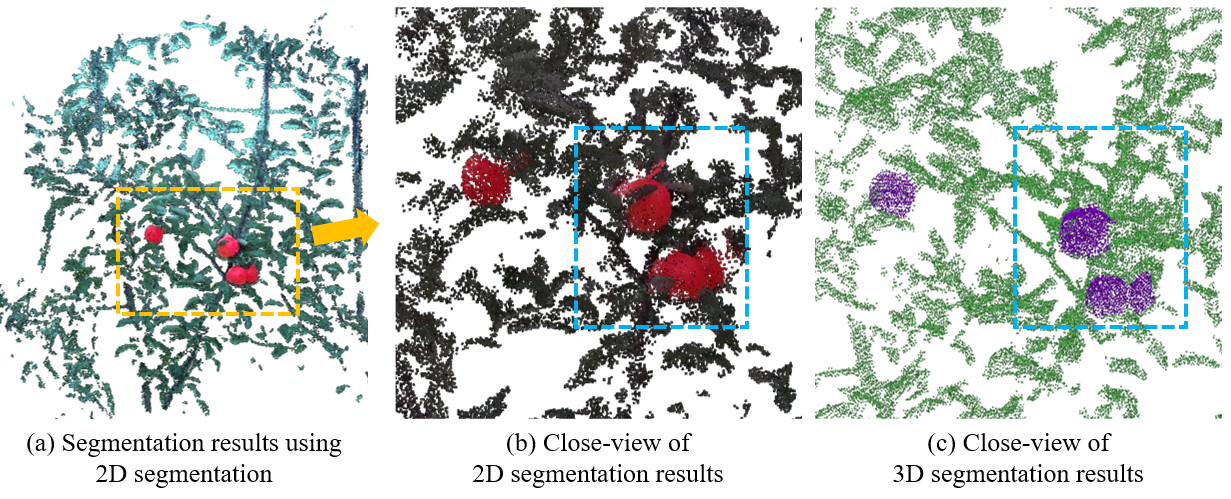}
\caption{The segmentation results of using the 2D segmentation method, (a) segmentation result of the complete scene; (b) details of the segmentation result; (c) details of the segmentation result using the proposed 3D segmentation method.}
\label{fig: rgbd segmentation}
\end{figure}

We further compare the segmentation accuracy of our proposed method and the SOTA 2D segmentation network Deeplab-v3+ $^{2}$. The training data of 2D segmentation is from the 2D RGB image of each fused data. We use 76 images to train the network, 38 images for validation during training, and the rest 38 for evaluation. Image augmentation such as color adjustment in HSV space, translation, and rotation is applied. The training resolution is 512 $\times$ 512. We used ResNet51 as the backbone and pre-trained models from Pascal VOC and Cityscapes datasets. The mIoU of the trained network on 2D segmentation is 0.8075. The point cloud colorized by using segmentation results from 2D segmentation methods is shown in Figure \ref{fig: rgbd segmentation}. 
\footnotetext[2]{https://github.com/VainF/DeepLabV3Plus-Pytorch}

It can be seen that 3D segmentation methods achieve better accuracy on fruit segmentation compared to 2D segmentation. As shown in Figure \ref{fig: rgbd segmentation}, laser beam divergence angle may cause fake points near the objects' edge. Due to lack of information from the point cloud, 2D segmentation methods have limited performance when dealing with this inherent defect of LiDAR data, while 3D segmentation capability in this condition. Meanwhile, 3D segmentation methods can also be easily integrated with SLAM, a common framework to obtain the point cloud models of large-scale scenes. 3D segmentation network can be directly applied to the colorized 3D point clouds. This means the segmentation task takes both geometry and color information into the feature extraction process, which can potentially compensate for the error caused by the noisy outlier points in 3D point clouds and by the Lidar-camera extrinsic calibration. However, 2D segmentation is conducted on RGB images initially and requires an accurate calibration between 2D to 3D data to achieve promising performance.

\section{Discussions}\label{section: discussion}
This study proposes a deep-learning-based point cloud segmentation network to perform semantic segmentation on fused colorized point cloud data acquired using the LiDAR-Camera sensor. An octree subdivision algorithm is used to subdivide the complete scene into multiple blocks to perform segmentation on a large number of points (100k - 200k points). The point cloud from each block is processed by the network and then added together to form the final segmentation results. Multi-sensor data fusion and network training under imbalance class distribution are two critical problems that need to be solved in point cloud segmentation network training. For multi-sensor data fusion, we analyze the architecture of the PointNet++ model and develop a later-fusion strategy to fuse color features into point cloud processing, which largely improves the segmentation accuracy of the model. For network training under imbalance class distribution, we utilize the under-sampling strategy in data pre-processing and WCE loss function during training, which secures the network training performance in dealing with the imbalance class training problem.

In the experiments, we analyze the effect of different sampling and grouping strategies in SA block within the pointNet++ architecture. Results show that the insufficient number of centroids and grouping of neighbour points may lead to network performance degeneration. In contrast, the excessive number of centroids and grouping of neighbour points can largely increase the computational requirements. After that, we evaluate the network performance by applying different fusion designs in network architecture. The results show that the proposed later-stage fusion can largely improve the network accuracy using fused data from LiDAR-Camera. Moreover, we analyze the imbalance class problem in network training. The under-sampling and WCE are utilized to enhance the training efficacy of the network. Overall, with proper design of the network architecture and training strategies, the proposed Fuse-PointNet++ model achieves 0.881 on the evaluation dataset and 0.862 on complete scene segmentation from the real orchard.

For comparison, we use the 3D segmentation network on fused data from an RGB-D camera. Results show that the accuracy of the network on the RGB-D camera shows a large reduction. The possible reason for this performance degeneration is the low accuracy of the point cloud data from the RGB-D camera. This result indicates that an accurate input point clout is essential to secure the performance of the point-based segmentation network. Furthermore, we compare the segmentation performance using the SOTA 2D segmentation method. Results show that 3D segmentation achieves a better score on accuracy. Meanwhile, since 3D segmentation methods can efficiently learn the geometrical features by the multi-scale grouping operations, our proposed Fuse-PointNet++ model has better capability to classify noise points near the edge of the objects.

\section{Conclusion}\label{section: conclusion}
In this work, a deep-learning-based 3D segmentation network is presented to perform fruit segmentation on a high-resolution point cloud. The proposed method can efficiently fuse the features from both objects' texture and geometrical appearances. Meanwhile, the proposed method utilized an under-sampling data pre-processing strategy and WCE loss function to improve the training performance when dealing with imbalance class problems. Our method is evaluated in both the evaluation dataset and complete scene segmentation, which are collected by using the LiDAR-Camera sensor in apple orchards. Overall, the proposed Fuse-PointNet++ network model achieved a mIoU of 0.881 on the evaluation dataset and 0.862 on complete scene segmentation. Experimental results show that our proposed method can perform accurate semantic segmentation of fruits in real orchards environments. Future work will investigate the feature fusion and network training strategies under the imbalance class condition to improve the segmentation accuracy on the point cloud.


\bibliographystyle{IEEEtran}
\bibliography{root}

\end{document}